\documentclass[letterpaper]{article} 
\usepackage{aaai2026}  
\usepackage{times}  
\usepackage{helvet}  
\usepackage{courier}  
\usepackage[hyphens]{url}  
\usepackage{graphicx} 
\usepackage{natbib}  
\usepackage{caption} 
\usepackage{algorithm}
\usepackage{algorithmic}

\usepackage{newfloat}
\usepackage{listings}
\DeclareCaptionStyle{ruled}{labelfont=normalfont,labelsep=colon,strut=off} 
\floatstyle{ruled}
\newfloat{listing}{tb}{lst}{}
\floatname{listing}{Listing}
\usepackage{amsmath}
\usepackage[capitalize]{cleveref}
\crefname{section}{Sec.}{Secs.}
\Crefname{section}{Section}{Sections}
\Crefname{table}{Table}{Tables}
\crefname{table}{Tab.}{Tabs.}
\usepackage{tabularx}
\usepackage{booktabs}
\usepackage{amssymb}
\usepackage{bbding}
\usepackage{pifont}
\usepackage{wasysym}
\usepackage{utfsym}
\usepackage{fontawesome}
\usepackage{array}
\usepackage{subfigure}
\usepackage{multirow}
\usepackage{makecell}
\usepackage{dsfont}

\newtheorem{theorem}{Theorem}
\newtheorem{lemma}{Lemma}

\newtheorem{remark}{Remark}

\title{Leveraging Dissimilarity Invariance as a Robust Anchor for Learning with Noisy Labels}

\author {
    Wenxiao Fan,
    Kan Li\textsuperscript{\footnotemark[2]}
}
\affiliations {
    School of Computer Science, Beijing Institute of Technology\\
    \{wenxiaofan, likan\}@bit.edu.cn
}

\usepackage{bibentry}

\begin{document}

\maketitle

\begin{abstract}
Deep learning models excel in visual recognition but suffer severe performance drops when training labels are corrupted by noise. Under label noise prior work cannot learn accurate similarities and thus misguide the learning process. In this paper, we uncover a complementary and novel phenomenon, Dissimilarity Invariance, whereby semantic dissimilarity between unrelated samples remains stable despite label noise. Leveraging this insight, we propose NegScale, a plug-and-play framework that shifts focus from fragile similarity to robust dissimilarity. NegScale integrates: (1) Structured Negative Orthogonality Penalty (SNOP), enforcing subspace orthogonality for unrelated samples; and (2) Dissimilarity-Calibrated Similarity Adjustment (DCSA), suppressing spurious similarity using dissimilarity anchors. We also give theoretical analysis that proves Dissimilarity Invariance and the effectiveness of NegScale. Empirical results demonstrate that NegScale consistently outperforms state-of-the-art baselines, establishing new benchmarks on CIFAR with synthetic noise and real-world datasets.
\end{abstract}


\section{Introduction}

\begin{figure}[t]
\centering
\includegraphics[width=1\columnwidth]{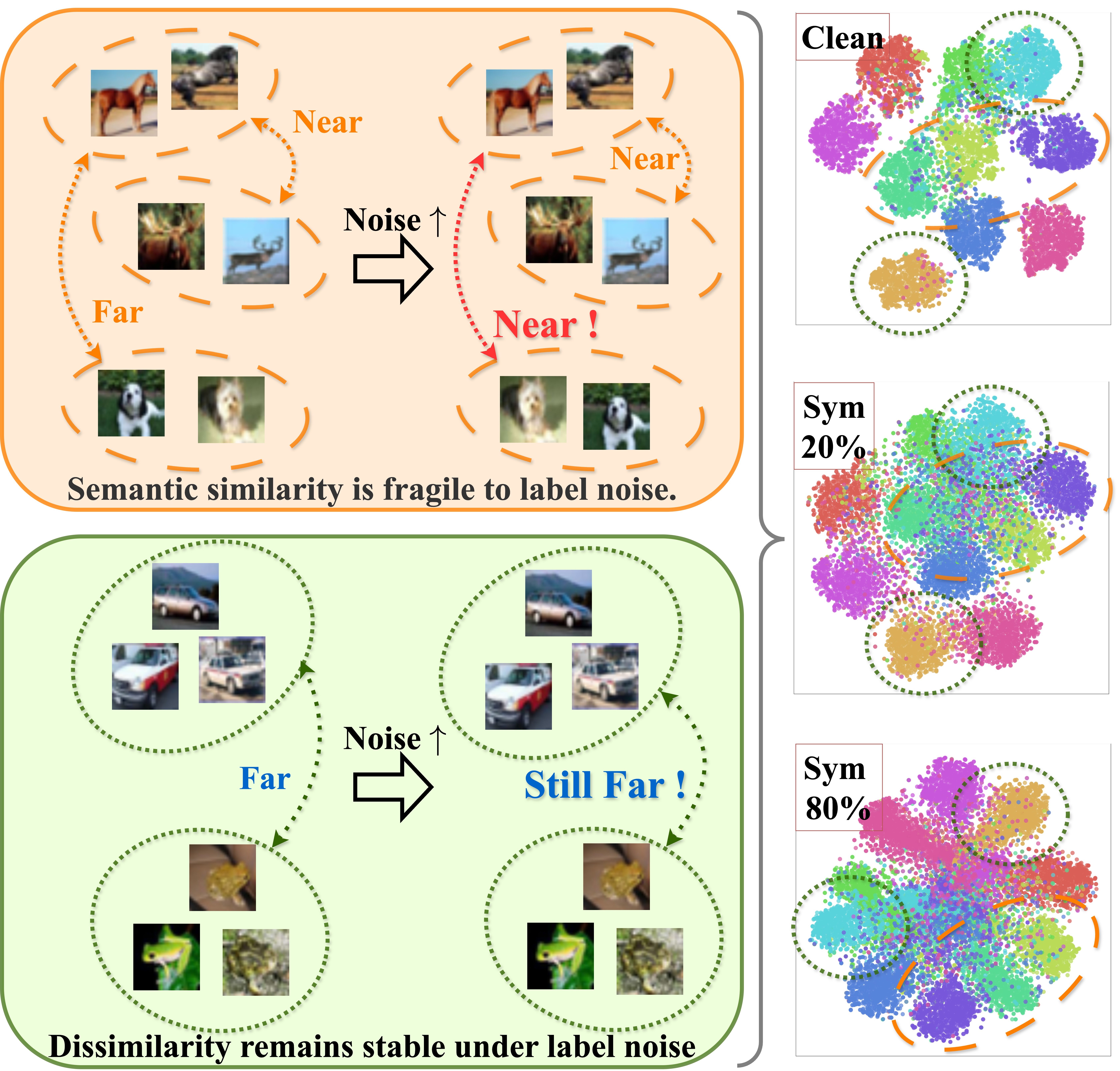} 
\caption{Illustration of Dissimilarity Invariance. Semantic similarity (e.g., horse vs. dog, as highlighted in the orange region above) is vulnerable to label noise, exhibiting significant variations under corruption. In contrast, inter-class dissimilarity (e.g., automobile vs. frog, shown in the green region below) remains largely stable.}
\label{fig:intro}
\end{figure}

Deep learning models have achieved remarkable success across a wide range of visual recognition tasks \cite{DBLP:journals/corr/abs-2004-10934,DBLP:conf/cvpr/MarriottR021}, but their performance degrades sharply when training labels are corrupted by noise. 
One underlying cause of this brittleness lies in how noisy labels distort the semantic structure of the data, especially the learned similarities between samples \cite{Chen2021TwoWD,DBLP:conf/aaai/FanL25}.
In practice, we find that similarity are inherently fragile under label noise (see \cref{fig:intro}). When samples presumed to belong to the same or semantically related classes are corrupted by incorrect labels, their pairwise similarities become severely distorted. Aligning the model according to these erroneous affinities not only degrades performance but also prevents the model from learning meaningful relationships. 

In contrast, we identify an underexplored phenomenon, named \textbf{Dissimilarity Invariance}, in which semantic similarity between unrelated samples (negative pairs)—referred to as dissimilarity—remains remarkably stable even as label noise increases. Specifically, representations of semantically unrelated classes maintain consistent dissimilarity levels across a wide range of noise rates. This observation suggests that, while label noise can severely distort similarity, it has far less impact on dissimilarity. As a result, models can reliably learn and leverage these robust dissimilarity patterns to improve performance under noisy supervision.

\begin{figure*}
    \centering
    \subfigure[Clean-Clean pairs ($y_i=\tilde{y}_i,y_j=\tilde{y}_j$)]
     { \begin{minipage}[t]{0.3\linewidth}
			\centering
			\includegraphics[width=1\linewidth]{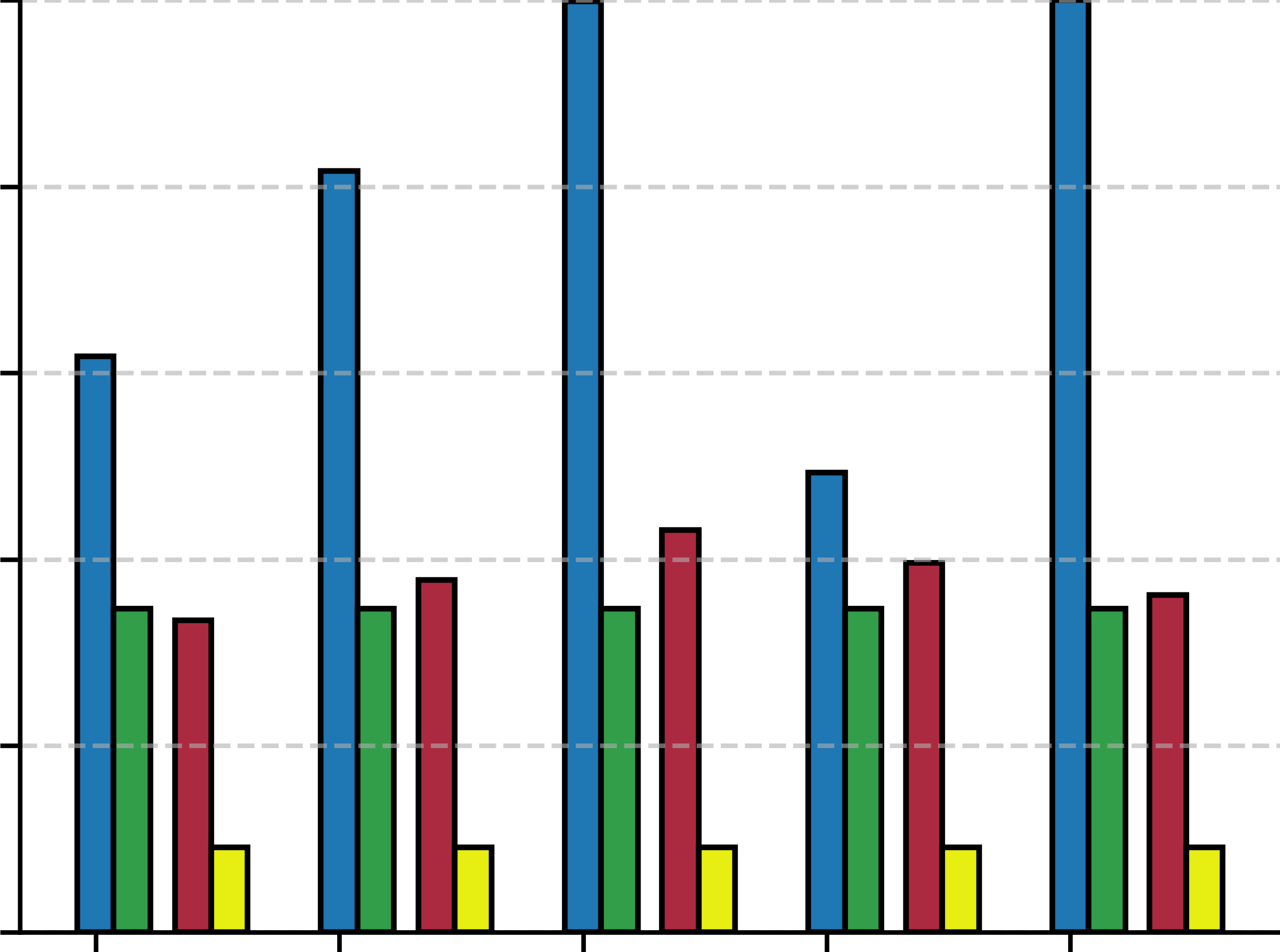}
			\label{fig:app_key_observation_a}
		\end{minipage}
		}
     \hfill
    \subfigure[Clean-Noisy pairs ($y_i=\tilde{y}_i,y_j\neq\tilde{y}_j$)]
     { \begin{minipage}[t]{0.3\linewidth}
			\centering
			\includegraphics[width=1\linewidth]{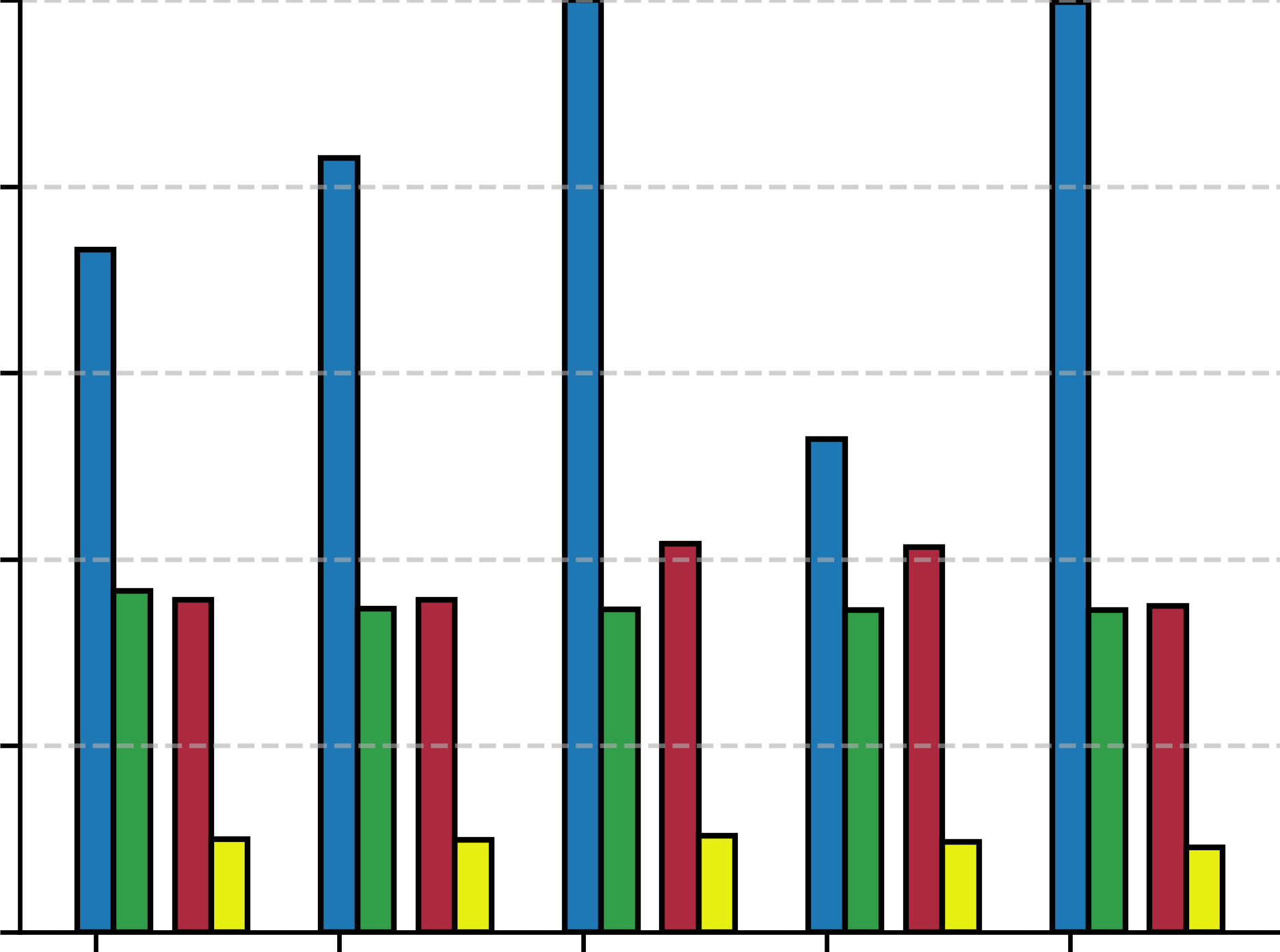}
			\label{fig:app_key_observation_b}
		\end{minipage}
		}
    \hfill
     \subfigure[Cross-noise pairs ($y_i=\tilde{y}_i=\tilde{y}_j\neq y_j$)]
     { \begin{minipage}[t]{0.3\linewidth}
			\centering
			\includegraphics[width=1\linewidth]{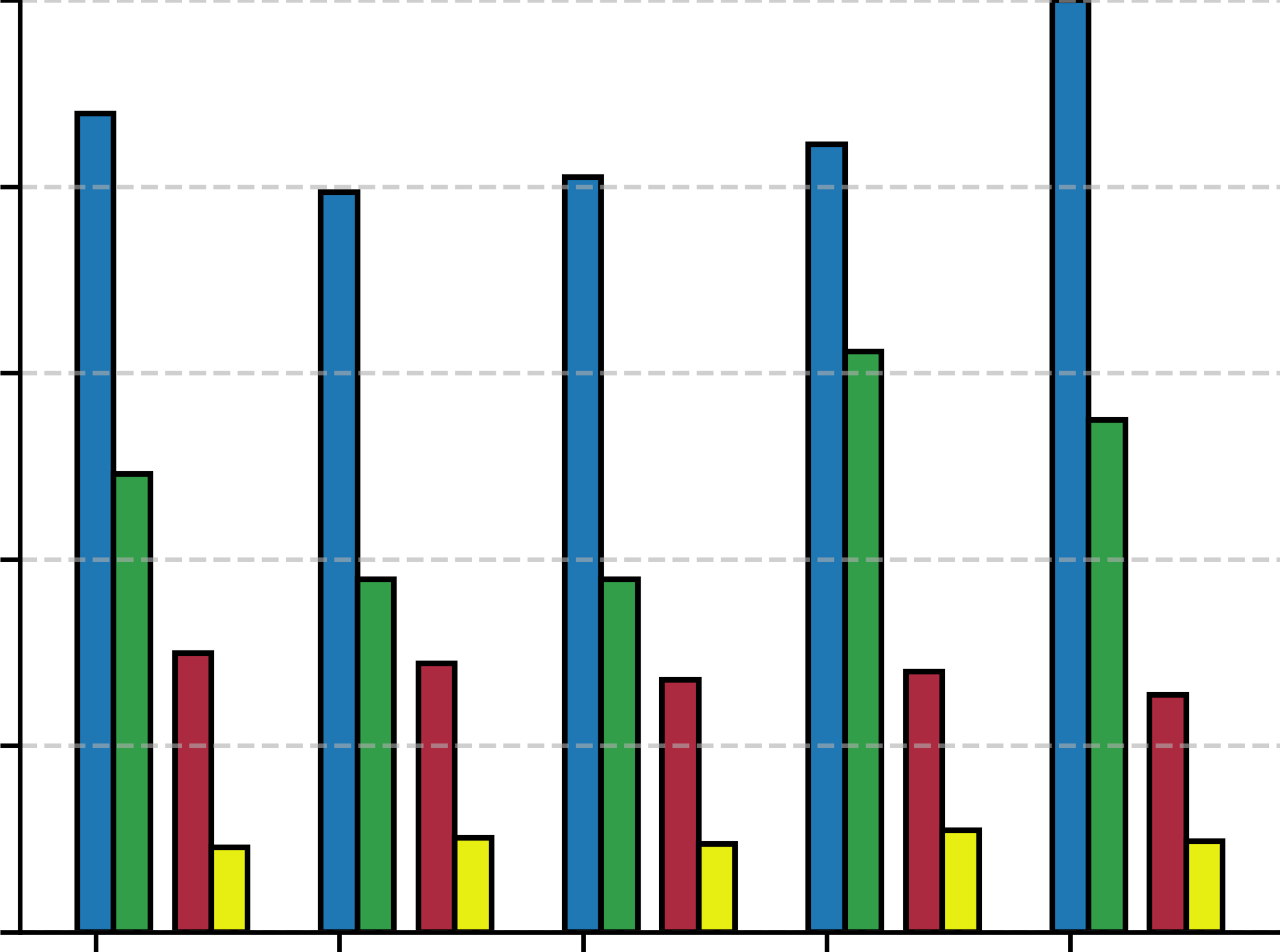}
			\label{fig:app_key_observation_c}
		\end{minipage}
		}
    \hfill
    \caption{Key observations of dissimilarity invariance under synthetic noises with CIFAR-10. Small / Large pLCA refers to semantically related and semantically unrelated sample pairs, respectively. Small / Large Clean indicates the similarity of the corresponding pairs when trained under clean labels. The y-axis represents the normalized similarity. The x-axis shows the noise rates, from left to right: [Sym 20\%, Sym 50\%, Sym 80\%, Ins 20\%, Ins 40\%]. The different colors correspond to the sets of \textcolor[HTML]{1f77b4}{\rule{7pt}{7pt}} {Small pLCA}, \textcolor[HTML]{329e49}{\rule{7pt}{7pt}} Small Clean, \textcolor[HTML]{ac2a40}{\rule{7pt}{7pt}} {Large pLCA}, and \textcolor[HTML]{e7ee11}{\rule{7pt}{7pt}} {Large Clean}.}
    \label{fig:app_key_observation}
\end{figure*}

Motivated by this observation, we shift focus from fragile similarity toward reliable dissimilarity as the robust anchor. 
We present Negative Scale (NegScale), a novel, plug-and-play framework designed to robustly model dissimilarity and suppress spurious similarity under label noise. NegScale comprises two modules: Structured Negative Orthogonality Penalty (SNOP), which enforces orthogonality among negative pairs within each minibatch to ensure that genuine negative pairs occupy distinct subspaces; and Dissimilarity-Calibrated Similarity Adjustment (DCSA), which leverages dissimilarity anchors to identify and down-weight misleading similarity, thereby preventing the model from internalizing false associations. We further provide a theoretical analysis to explain the underlying cause of Dissimilarity Invariance and to justify the effectiveness of NegScale. Requiring no external data or pre-training, NegScale can be seamlessly integrated into existing training pipelines. Empirical evaluations on both synthetic and real-world noisy-label benchmarks confirm that our dissimilarity-centric approach consistently outperforms SOTA methods.
In a nutshull, our contributions are as follows:
\begin{itemize}
    \item We observe a key phenomenon, Dissimilarity Invariance, wherein the similarity between semantically unrelated samples remains notably more robust under label noise.
    \item We propose a novel framework, called NegScale, which enables the model to capture more accurate dissimilarity relationships while simultaneously suppressing spurious similarity via dissimilarity anchors, thereby enhancing its robustness.
    \item Experimental results show that our method advances state-of-the-art results on CIFAR with synthetic label noise, as well as on real-world noisy datasets.
\end{itemize}


\section{Releated Work}
Numerous research have recently addressed the issue of learning from noisy labels. Depending on how they approach handling noisy datasets, we categorize the current algorithms into three main groups and one subcategory.

\textit{Loss Correction.}
In order to reduce the negative impact of noisy labels, loss correction methods adjust the loss of all training samples before updating the parameters of the model \cite{DBLP:conf/icml/ArazoOAOM19,DBLP:conf/iccv/LiYSCLL17,Song2019}.
%
%
%
The estimated noise transition matrix or other approaches are used by the loss correction methods to modify the loss of all training samples, which is then used to update the network parameters .
However, the noise transition matrix's parameters are highly challenging to estimate, and comparable adjustments made to all samples invariably suffer from the accumulated incorrect correction. This can have an enormous impact on the model's final performance \cite{Jiang2017,Han2018c,DBLP:journals/corr/abs-2007-08199}.
\textit{Sample Selection.}
To avoid the false correction, many studies use the sample selection to improve the performance of the model \cite{Chen2019,Song2019,Han2018c}.
ANNE \cite{DBLP:journals/pr/CordeiroC25} uses integrates loss-based sampling with the feature-based sampling methods FINE and Adaptive KNN.
However, 
this family of methods discard a very large number of samples and select only a portion of samples for learning, which are easy to result in knowledge waste.
\textit{Semi-supervised Learning.}
In terms of the problems with the previous methods such as false correction and knowledge waste, Semi-supervised Learning is proposed and has achieved excellent results in recent years \cite{DBLP:journals/corr/abs-2007-08199,Chen2021TwoWD,DBLP:conf/aaai/FanL25}. The core idea of Semi-supervised Learning is treating the possibly noisy samples as unlabeled, whereas the rest samples as labeled. 
However, this family of methods requires careful setting of hyperparameters, which indirectly increases the complexity of the calculation.

\section{Methodology}

\begin{figure*}
    \centering
    \subfigure[Clean-Clean pairs ($y_i=\tilde{y}_i,y_j=\tilde{y}_j$)]
     { \begin{minipage}[t]{0.3\linewidth}
			\centering
			\includegraphics[width=1\linewidth]{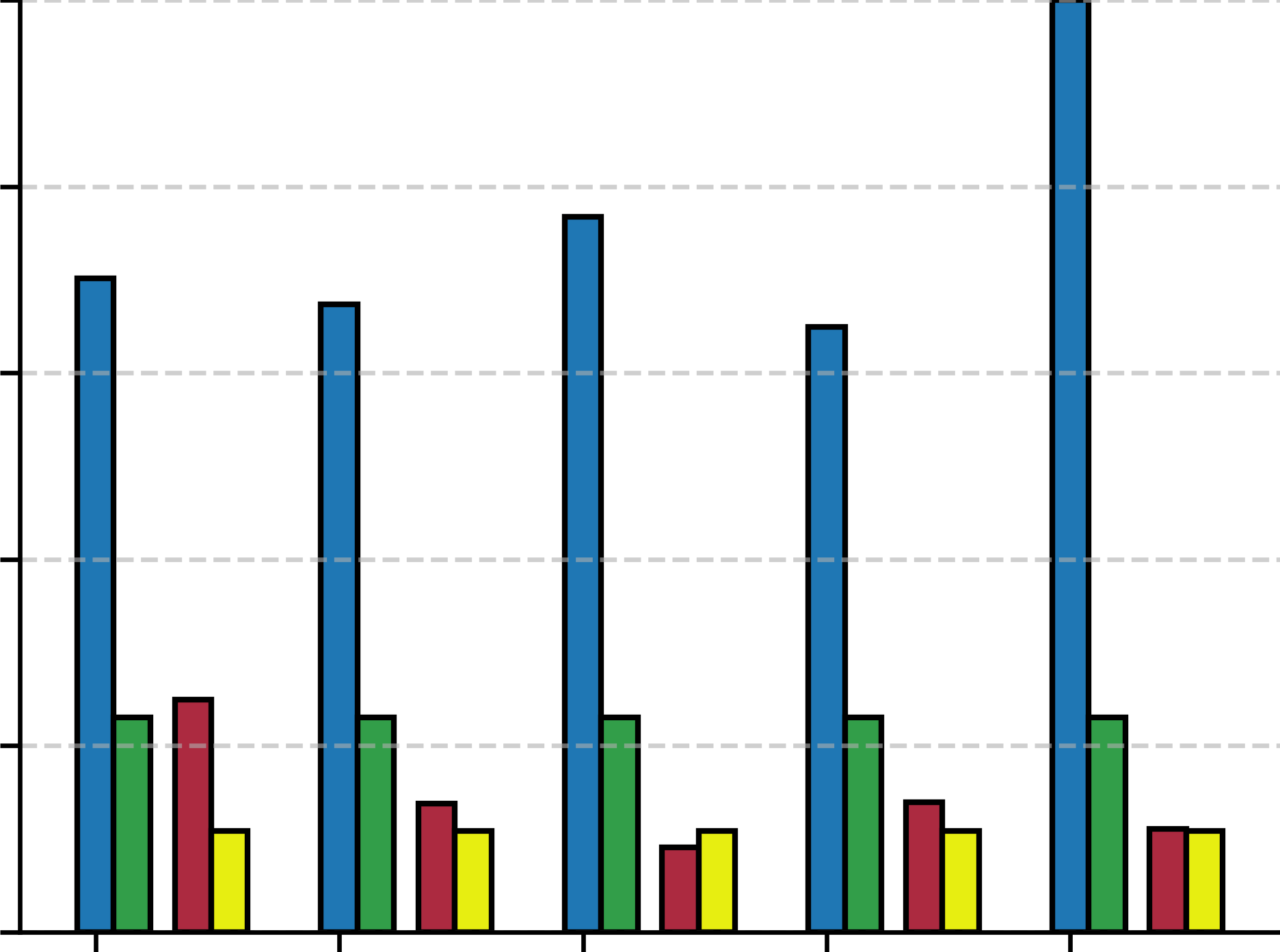}
			\label{fig:key_observation_a}
		\end{minipage}
		}
     \hfill
    \subfigure[Clean-Noisy pairs ($y_i=\tilde{y}_i,y_j\neq\tilde{y}_j$)]
     { \begin{minipage}[t]{0.3\linewidth}
			\centering
			\includegraphics[width=1\linewidth]{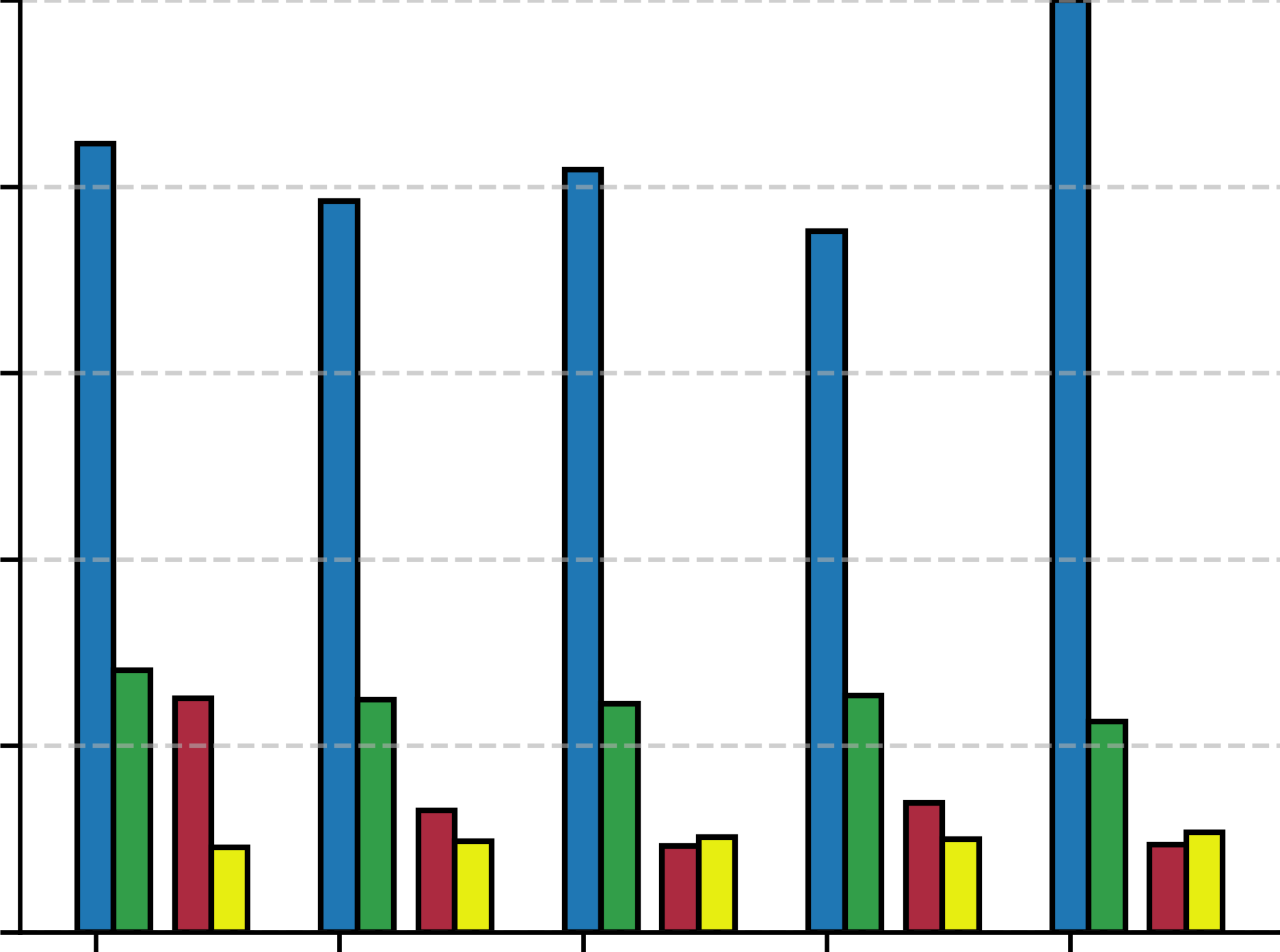}
			\label{fig:key_observation_b}
		\end{minipage}
		}
    \hfill
     \subfigure[Cross-noise pairs ($y_i=\tilde{y}_i=\tilde{y}_j\neq y_j$)]
     { \begin{minipage}[t]{0.3\linewidth}
			\centering
			\includegraphics[width=1\linewidth]{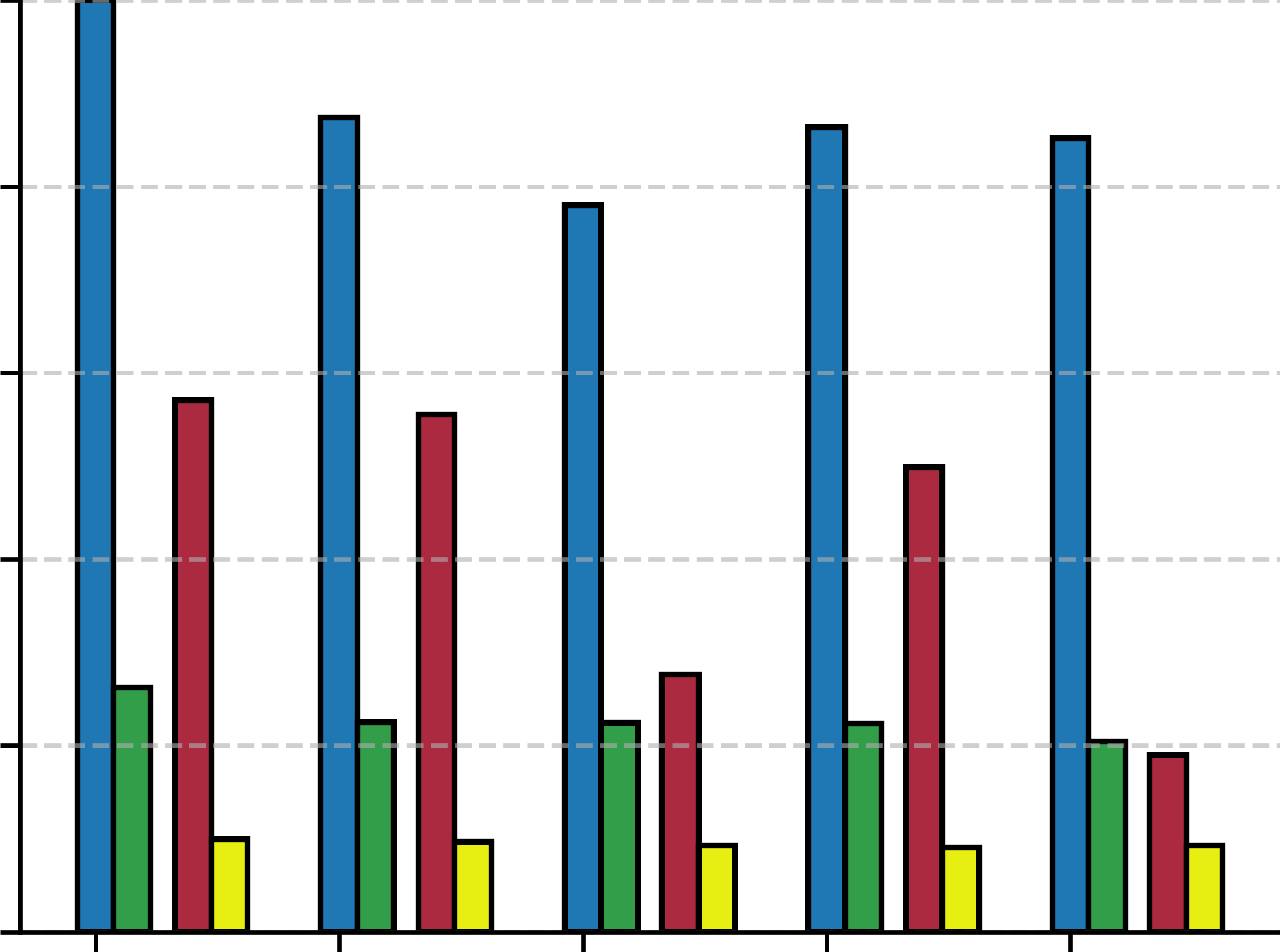}
			\label{fig:key_observation_c}
		\end{minipage}
		}
    \hfill
    \caption{Key observations of dissimilarity invariance under various noise types of real world noisy dataset CIFAR-10N. Small / Large pLCA refers to semantically related and semantically unrelated sample pairs, respectively. Small / Large Clean indicates the similarity of the corresponding pairs when trained under clean labels. The x-axis shows the noise types, from left to right: [Aggre, Rand1, Rand2, Rand3, Worst]. The legend is consistent with \cref{fig:app_key_observation}.}
    \label{fig:key_observation}
\end{figure*}

\begin{table}[htbp!]
  \small
  \centering
    \begin{tabular}{lcccc}
    \toprule
    \textbf{Noise Type} &\textbf{Acc}   & \makecell[c]{\textbf{Same}\\\textbf{Class}} & \makecell[c]{\textbf{Small} \\\textbf{pLCA}} & \makecell[c]{\textbf{Large}\\\textbf{pLCA}} \\
    \midrule
    Clean & 95.72 & 0.94  & 0.48  & 0.44 \\
    \midrule
    Sym 20\% & 87.32 & 0.80  & 0.54  & 0.48 \\
    Sym 50\% & 80.04 & 0.74  & 0.55  & 0.48 \\
    Sym 80\% & 56.89 & 0.63  & 0.58  & 0.49 \\
    \midrule
    Aggre 8\% & 92.16 & 0.89  & 0.68  & 0.49 \\
    Rand1 17\% & 89.13 & 0.82  & 0.66  & 0.45 \\
    Rand2 17\% & 89.14 & 0.84  & 0.67  & 0.43 \\
    Rand3 17\% & 89.39 & 0.83  & 0.65  & 0.45 \\
    Worst 40\% & 83.31 & 0.82  & 0.73  & 0.44 \\
    \midrule
    Mean  & -     & 0.80  & 0.60  & 0.46 \\
    $\Delta$ With Clean   & -     & -0.14 & 0.12  & 0.03 \\
    \bottomrule
    \end{tabular}%
      \caption{Accuracy and similarity variations of ResNet-18 trained with Cross-Entropy on CIFAR-10 under various noise.
      Line 3-7 correspond to synthetic noise settings, while Line 8-12 represent real-world noise from CIFAR-10N.
      }
  \label{tab:key_observation}%
\end{table}%

\noindent\textbf{Preliminary.}
Let $\mathcal{D} = \{(x_i, \tilde{y}_i)\}_{i=1}^N$ denote a training set with noisy labels, where $x_i \in \mathcal{X}$ is the input sample and $\tilde{y}_i \in \mathcal{Y} = \{1, \dots, C\}$ is its (possibly corrupted) label, the ground-truth label is $y_i$, $N$ is the number of samples, $C$ is the number of classes and $\mathcal{B}$ is the batch. We denote the feature extractor by $f(\cdot; \theta)$ and the classifier head by $g_c(\cdot)$. The full model prediction is given by $g_c(f(x_i))$. The noise rate is $\tau$.
We define the cosine similarity between two feature vectors $f_i = f(x_i)$ and $f_j = f(x_j)$ as:
$s_{ij} = \frac{f_i^\top f_j}{\|f_i\| \cdot \|f_j\|}$.
Our goal is to train a robust model that learns clean feature representations even under such label corruption.

\noindent\textbf{How to Determine the Semantic Relatedness between Two Samples and Identify Negative Pairs?}
We use the pair Lowest Common Ancestor (pLCA) distance \cite{DBLP:conf/cvpr/BertinettoMTSL20,DBLP:conf/icml/ShiGTCLVFFK24}, 
which measures the semantic distance 
between two classes based on class taxonomy, as demonstrated in \cref{eq:pLCA}. A lower pLCA score indicates a closer semantic relationship between two classes $(y,\hat{y})$. 
\begin{equation}
    \label{eq:pLCA}
    D_{pLCA} (y,\hat{y}) \overset{\underset{\mathrm{def}}{}}{=}  |h(y)-h(N_{L} (y,\hat{y}))| + |h({\hat{y}})-h(N_{L} (y,\hat{y}))|
\end{equation}
where $h$ represents the tree depth of a node and $N_{L}$ is the lowest common ancestor class node for the label within the hierarchy. The predefined taxonomic hierarchy of pLCA distance is shown in the Appendix.
We can leverage the pairwise label pLCA distance to determine the semantic relatedness between two samples. 
Samples are considered semantically related if their pLCA distance is below threshold $\eta_{\min}$, and semantically unrelated if exceeding threshold $\eta_{\max}$, $\eta_{\min}\leq \eta_{\max}$. Within each training epoch, we first freeze the model parameters to compute inter-sample similarities and pLCA distances, get and normalize the set of semantically related samples $\mathcal{P}_{\text{sim}}=\{(i,j)|0<D_{pLCA}(\tilde{y}_i,\tilde{y}_j)\leq\eta_{\min}\}$ and unrelated samples $\mathcal{P}_{\text{neg}}=\{(i,j)|D_{pLCA}(\tilde{y}_i,\tilde{y}_j)>\eta_{\max}\}$, then unfreeze the parameters to perform gradient-based optimization of the loss function.

\begin{table*}[htbp]
  \centering
  \small
    \begin{tabular}{l|ccc|c|c|ccc|c|c}
    \toprule
    \textbf{Dataset} &      \multicolumn{5}{c|}{\textbf{CIFAR-10}} & \multicolumn{5}{c}{\textbf{CIFAR-100}} \\
    \midrule
    \textbf{Noise Type} & \multicolumn{3}{c|}{\textbf{Sym}}  & \multicolumn{1}{c|}{\textbf{Pair}} &  \multicolumn{1}{c|}{\textbf{Ins}}  & \multicolumn{3}{c|}{\textbf{Sym}}  & \multicolumn{1}{c|}{\textbf{Pair}}  & \multicolumn{1}{c}{\textbf{Ins}}   \\
    \midrule
    \textbf{Method / Noise Rate} & \textbf{20\%} & \textbf{50\%}& \multicolumn{1}{c|}{\textbf{80\%}}& \multicolumn{1}{c|}{\textbf{40\%}}  & \multicolumn{1}{c|}{\textbf{40\%}}  & \textbf{20\%} & \textbf{50\%}& \multicolumn{1}{c|}{\textbf{80\%}}& \multicolumn{1}{c|}{\textbf{40\%}} & \multicolumn{1}{c}{\textbf{40\%}}    \\
    \midrule
    Co-teaching \cite{Han2018c} & 88.2  & 50.7  & 21.1 & 55.3 & 59.5  & 58.5  & 33.0  & 5.8 & 39.2 & 40.7 \\
     

    DivideMix \cite{DBLP:conf/iclr/LiSH20}& 95.7 &	94.4&	92.9	&	92.1	&	95.1& 76.9	&74.2&	59.6&	52.3&	76.1\\
    Co-learning \cite{DBLP:conf/mm/TanXWL21} & 91.8  & 79.3  & 37.0 & 66.3 &  78.9  & 70.3  & 63.9  & 38.9 & 49.1 & 62.9  \\
     SELC+ \cite{Lu2022SELCSL} &{{94.9}}& 87.2 &{78.6} & 88.1 & 84.2 &{76.4}& 62.4 &{37.2} & 45.2 & 44.3 \\

    RoLR \cite{Chen2021TwoWD} & 96.4&	95.7&	94.2&	92.8 & 	93.7&  78.6&	74.6&	66.2& 76.1 & 77.2  \\
    RankMatch \cite{DBLP:conf/iccv/Zhang0FLCLL23} & 96.4 &	95.4&	94.2&  94.4 & 93.8  & 79.3&	77.6&	67.2&  75.8& 76.5  \\
    CrossSplit \cite{DBLP:conf/icml/KimBZL23} & 96.9	&96.3	&95.4			&96.0 & 95.8 &79.9& 75.7 & 64.6&76.8 & 79.2  \\
    DMLP (Naive) \cite{DBLP:conf/cvpr/TuZLLLWWZ23}& 94.2&	94.0 &93.2&93.9& 93.2& 72.3& 70.1	&	63.2& 71.8 & 72.2\\
    DMLP (DivideMix) \cite{DBLP:conf/cvpr/TuZLLLWWZ23} &96.2	&95.6	&94.3&	95.0  &95.4 & 79.4	&76.1&	68.5& 76.4 &  78.9 \\
    CCL \cite{DBLP:conf/aaai/FanL25} &  {97.0}&	{96.5}&	{94.6}&  {96.1}	&	{96.2} & {79.5}	&{77.4}&	{70.3} & {77.2}	 &	{80.0} \\
    ANNE \cite{DBLP:journals/pr/CordeiroC25} &96.9 & 96.2 & 95.3 & 95.7 &96.2 & 80.4 & 78.1 & \textbf{73.0} & 66.4 & 78.4 \\

    \midrule
    RoLR + NegScale (Ours) & \textbf{97.2} & \textbf{96.6} & \textbf{95.6} & \textbf{96.3} & \textbf{96.5} & \textbf{80.9} & \textbf{78.7} & {70.8} & \textbf{77.8} & \textbf{80.4} \\
    \bottomrule
    \end{tabular}%
    \caption{Comparison with state-of-the-art methods on CIFAR-10/100 datasets under various types of noise. 
    The results of other methods are from the published results of corresponding papers.
    The best results are indicated in bold.
     }
  \label{tab:main_1}%
\end{table*}%

\subsection{Key Observations of Dissimilarity Invariance}
\label{sec:key_ob}

While LNL methods can extract robust features under noisy labels, they often neglect structural relationships among samples, causing learned similarities to become fragile and susceptible to error amplification. To investigate this issue, we examine sample-pair similarities across different noise types. Our analysis reveals a clear contrast: similarities between semantically related samples (small pLCA) vary significantly as noise increases, whereas those between unrelated samples (large pLCA)—which we refer to as dissimilarity—remain remarkably stable. Negative pairs consistently maintain low cosine similarity with minimal variance across noise rates (\cref{tab:key_observation}). We term this effect \textbf{Dissimilarity Invariance}, emphasizing that label noise primarily distorts positive relations while leaving dissimilar pairs largely unaffected.

To further investigate Dissimilarity Invariance, we categorize sample pairs based on the degree of label corruption and analyze how their similarity deviates, as shown in \cref{fig:key_observation,fig:app_key_observation}. Specifically, we divide each sample pair $(i,j)$ into three categories:
(1) Clean-Clean pairs: both samples have clean labels, as $y_i=\tilde{y}_i,y_j=\tilde{y}_j$;
(2) Clean-Noisy pairs: one sample has a noisy label, as $y_i=\tilde{y}_i,y_j\neq\tilde{y}_j$;
(3) Cross-noise pairs: one sample has a noisy label, and that noisy label coincides with the clean label of the other sample, as $y_i=\tilde{y}_i=\tilde{y}_j,y_j\neq\tilde{y}_j$.
\cref{fig:key_observation,fig:app_key_observation} further support three key conclusions: (1) Dissimilarity Invariance holds universally, regardless of whether the samples are affected by noise, as shown in \cref{fig:key_observation_a,fig:key_observation_b,fig:app_key_observation_a,fig:app_key_observation_b}.
(2) The effect of Dissimilarity Invariance is more pronounced under real-world noise compared to synthetic noise.
(3) Mislabeled samples significantly disrupt similarity within semantically related pairs, but their impact on negative pairs remains notably smaller, as illustrated in \cref{fig:key_observation_c,fig:app_key_observation_c}.




\subsection{NegScale: Learning with Stable Dissimilarity}

Building on our core observation that dissimilarity between negative pairs remains stable even under label noise, we derive two key insights: (1) accurate dissimilarity can still be learned despite noisy supervision; and (2) such reliable dissimilarity can be leveraged to rectify corrupted similarity signals. Guided by these insights, we propose NegScale, a robust learning framework composed of two complementary modules: SNOP, which enforces structured orthogonality among negative pairs, and DCSA, which calibrates noisy similarities using dissimilarity-aware adjustments.

\noindent\textbf{Structured Negative Orthogonality Penalty. }
SNOP enforces orthogonality among negative pairs at the batch level, encouraging their separation in the representation space. The underlying idea is that dissimilar samples should lie in orthogonal directions, reflecting semantic independence—a principle shown to improve robustness in prior work \cite{DBLP:conf/aaai/0003ZHL0W22,DBLP:journals/kbs/YuanY22}.
We detail the formulation of SNOP below.
First, we define the difference vector set as:
\begin{equation}
    F_{\text{neg}} = \left\{ \frac{f_i - f_j}{\|f_i - f_j\|_2} \,|\, (i, j) \in \mathcal{P}_{\text{neg}}\right\}
\end{equation}
where each column corresponds to a normalized difference between a dissimilar pair. Intuitively, if all these directions are orthogonal, then $F_{\text{neg}} F_{\text{neg}}^\top$ should approximate the identity matrix $I$. So we define the global orthogonality loss as: 
\begin{equation}
    \mathcal{L}_{\text{global}} = \|F_{\text{neg}} F_{\text{neg}}^\top - I\|_F^2
\end{equation}
While $F_{\text{neg}} F_{\text{neg}}^\top$ captures global repulsion structure, it may overlook local misalignments caused by particularly confusing or uncertain samples. To address this, we introduce a confidence-weighted local orthogonality loss that directly penalizes inner product alignment between negative pairs:
\begin{equation}
   \mathcal{L}_{\text{local}} =  { \sum_{(i,j)\in \mathcal{P}_{\text{neg}}} w_{ij} \cdot (f_i^\top f_j)^2}
\end{equation}
where the weight $w_{ij}$ is defined based on per-sample confidence as:
\begin{equation}
    w_{ij} = (1-c_i)\cdot(1-c_j)
\end{equation}
where $c_i$ is the confidence of sample $i$'s label prediction, computed as the softmax probability of its predicted class. 
This design assigns higher penalties to pairs involving low-confidence (potentially noisy) samples, enforcing stronger dissimilarity. In contrast, high-confidence pairs are deemed more reliable and penalized less.
The full SNOP loss is a combination of the global and local penalties:
\begin{equation}
\label{eq:snop}
\mathcal{L}_{\text{SNOP}} =
{\|F_{\text{neg}} F_{\text{neg}}^\top - I\|_F^2}
+
{\sum_{(i,j)\in \mathcal{P}_{\text{neg}}} w_{ij} \cdot (f_i^\top f_j)^2}
\end{equation}



\noindent
\textbf{Dissimilarity-Calibrated Similarity Adjustment.}
Under label noise, many such pairs are mislabeled, leading to spurious similarity (\cref{fig:app_key_observation,fig:key_observation}).
To address this, DCSA imposes soft upper bounds on similarity using dissimilarity, preventing mismatched pairs from becoming overly similar.
For each pair $(i,j) \in \mathcal{P}_{\text{sim}}$, we assess their proximity to dissimilar samples by computing a adaptive calibration factor:

\begin{equation}
\delta_{ij} = \frac{1}{2}\max_{(p,q) \in \mathcal{P}_{\text{neg}}} (f_i^\top f_p + f_j^\top f_q),
\end{equation}

This reflects how much $i$ and $j$ are entangled with unrelated samples. 
A large $\delta_{ij}$ suggests that the pair may be semantically ambiguous or mislabeled, and thus their similarity should be restricted.
To prevent the model from learning overly confident similarity on such risky pairs, we define the DCSA loss as:
\begin{equation}
  \label{eq:dcsa}
\mathcal{L}_{\text{DCSA}} = \sum_{(i,j) \in \mathcal{P}_{\text{sim}}}
\left[ \max(0, s_{ij}- (1-\delta_{ij})) \right]^2
\end{equation}
\cref{eq:dcsa} penalizes positive pairs whose similarity exceeds a soft upper bound $(1-\delta_{ij})$, which adapts based on their proximity to negative samples. 
DCSA acts as a dissimilarity-aware gate that prevents overfitting to noisy positives. 

\noindent
\textbf{Final Objective.}  
The total training loss is composed as:
\begin{equation}
  \label{eq:loss}
\mathcal{L}_{\text{total}} = \mathcal{L}_{\text{org}} + \lambda \mathcal{L}_{\text{SNOP}} + \mu \mathcal{L}_{\text{DCSA}}
\end{equation}
The coefficients $\lambda$ and $\mu$ balance dissimilarity enforcement and noisy similarity suppression. As our method requires no additional information, it can serve as a plug-in by replacing $\mathcal{L}_{\text{org}}$ with the desired loss in existing methods.


\subsection{Theoretical Analysis}

\noindent\textbf{Why Negative-Pair Similarity Remains Invariant.}
We show that, under \emph{symmetric} label noise and vanilla cross-entropy SGD, the expected variation in the similarity of negative pairs is smaller than that of semantically related pairs via first-order change in cosine similarity.

Let \(f_i(t)\in\mathbb R^d\) be the \(\ell_2\)-normalized feature of sample \(x_i\) at SGD iteration \(t\). The linear classifier is \(W=[w_1,\dots,w_C]\in\mathbb R^{d\times C}\); each logit is   $p_i^{(k)} = w_k^\top f_i,\quad k=1,\dots,C$.
From the nature of semantic similarity, we know that semantically related pairs ${(i,j)}\in \mathcal{P}_{\text{sim}}$ tend to have a higher overlap in their logits ($z_i^{(k)}\approx z_j^{(k)}$ for most $k$), while semantically unrelated pairs ${(i',j')}\in \mathcal{P}_{\text{neg}}$ exhibit minimal overlap. In other words, the Kendall tau distance \cite{665905b2-6123-3642-832e-05dbc1f48979} between the logits of semantically similar pairs $\pi(i,j)$ is typically smaller than that of negative pairs $\pi(i',j')$. 
A single (unnormalized) gradient step on \(f_i\) is  
  \begin{equation}
    \begin{aligned}
        f_i (t+1) &= f_i(t) - \gamma\,g_i\\
    g_i = \nabla_{f_i}\mathcal L_i &= \sum_{k=1}^C p_i^{(k)}\,w_k \;-\; w_{\tilde y_i}
    \end{aligned}
  \end{equation}
where $\gamma$ is the learning rate. After renormalization \(\|f_i(t+1)\|=1\), the first-order change in feature is \(\Delta f_i=-\gamma\,g_i+O(\gamma^2)\). For a pair \((i,j)\) with \(y_i\neq y_j\), define
$s_{ij}(t) = f_i(t)^\top f_j(t)$.
After one gradient step (dropping \(O(\gamma^2)\) and renormalization effects), the updated similarity is
\begin{equation}
\begin{aligned}
s_{ij}(t+1)
&=  f_i(t+1)^\top  f_j(t+1)\\
&= \bigl(f_i - \gamma g_i\bigr)^\top\bigl(f_j - \gamma g_j\bigr)\\
&= s_{ij} - \gamma\bigl\langle g_i,\,f_j\bigr\rangle
           - \gamma\bigl\langle f_i,\,g_j\bigr\rangle
           + O(\gamma^2),
\end{aligned}
\end{equation}
so the first-order increment is
\begin{equation}
|\Delta s_{ij}|
=|s_{ij}(t+1)-s_{ij}(t)|
\approx |\gamma\,\Bigl[\langle g_i,\,f_j\rangle + \langle f_i,\,g_j\rangle\Bigr]|
\end{equation}
Without loss of generality, we first examine $\gamma\bigl\langle g_i,\,f_j\bigr\rangle$:
\begin{equation}
    \begin{aligned}
        \gamma\bigl\langle g_i,\,f_j\bigr\rangle &= \gamma\big[(\sum_{k=1}^C p_i^{(k)}\,w_k \;-\; w_{\tilde y_i}) \cdot f_j \big] \\
        &=\gamma\big[\sum_{k=1}^C p_i^{(k)}\,w_k f_j \;-\; w_{\tilde y_i} f_j \big]\\
        &=\gamma\sum_{k=1}^C p_i^{(k)}p_j^{(k)}-p_j^{(\tilde{y}_i)}\approx\gamma\sum_{k=1}^C p_i^{(k)}p_j^{(k)}
    \end{aligned}
\end{equation}
We drop $p_j^{(\tilde{y}_i)}$ because, for a model with reasonable discriminative ability, this value tends to be small when \(y_i\neq y_j\), and can therefore be safely ignored. A similar conclusion also holds for $\gamma\bigl\langle f_i,\,g_j\bigr\rangle$.
So for $(i,j)\in \mathcal{P}_{\text{sim}}$ and $(i',j')\in \mathcal{P}_{\text{neg}}$:
\begin{equation}
\begin{aligned}
    \label{eq:delta_s_ij}
|\Delta s_{ij}|-|\Delta s_{i'j'}| &= |2\gamma\Bigl[\sum_{k=1}^C p_i^{(k)}p_j^{(k)} - \sum_{k=1}^C p_{i'}^{(k)}p_{j'}^{(k)}\Bigr]|\\
&=|2\gamma\sum_{k=1}^C \Bigl[p_i^{(k)}p_j^{(k)}-p_{i'}^{(k)}p_{j'}^{(k)}\Bigr]|
\end{aligned}
\end{equation}
From the benigning, we know that the Kendall tau distance of 
$(i,j)$ is greater than that of $(i',j')$. According to the rearrangement inequality \cite{Cvetkovski2012}, this implies that \cref{eq:delta_s_ij} $>0$. This result indicates that, during each update, the similarity changes more significantly for semantically similar pairs than for semantically dissimilar ones, directly supporting the validity of Dissimilarity Invariance.

\noindent\textbf{Why NegScale is Effective.}
We theoretically show that NegScale mitigate noise-induced feature perturbations and margin erosion. Proofs appear in the appendix.

\begin{lemma}[Feature Perturbation Bound under NegScale]
\label{lem:feature_perturbation_bound}
\noindent Let $f_i$ and $\tilde{f}_i$ be the feature representations learned under clean and noisy labels respectively.  
Suppose the feature extractor $f$ is trained with NegScale.
Assume each of $\mathcal{L}_{\text{SNOP}}$ and $\mathcal{L}_{\text{DCSA}}$ is minimized to at most $\delta_{\text{SNOP}}, \delta_{\text{DCSA}}$ respectively, and that the gradient norm of cross-entropy loss on noisy labels is bounded by $\|\nabla \mathcal{L}_{\text{CE}}\| \le G$.
Then the feature perturbation due to noisy labels is upper bounded as:
\begin{equation}
\|\tilde{f}_i - f_i\| \le \frac{G \cdot \tau}{\lambda \cdot \delta_{\text{SNOP}} + \mu \cdot \delta_{\text{DCSA}}}
\end{equation}
\end{lemma}

\begin{remark}
This lemma quantifies how much the learned feature \( \tilde{f}_i \) deviates from its clean counterpart \( f_i \) under noisy supervision. The bound highlights that this deviation increases linearly with the noise rate \( \tau \) but is effectively suppressed by the regularization strengths \( \lambda \) and \( \mu \). As a result, incorporating SNOP and DCSA reduces representation instability caused by corrupted labels.
\end{remark}

\begin{table*}[htbp]
  \centering
    \small
    \begin{tabular}{ll|ccccc|c}
    \toprule
    \multicolumn{2}{l}{\textbf{Dataset}}  \vline & \multicolumn{5}{c}{\textbf{CIFAR-10N}} \vline& \multicolumn{1}{l}{\textbf{CIFAR-100N}} \\
    \multirow{2}{*}{\textbf{Method}}  & \textbf{Noise Type} & \textbf{Aggre} & \textbf{Rand1} & \textbf{Rand2} & \textbf{Rand3} & \textbf{Worst} & \textbf{Fine}   \\
          & \textbf{Noise Rate} & \textbf{9.0}\% & \textbf{17.2\%} &\textbf{18.12\%} &\textbf{17.64}\%& \textbf{40.2\%}& \textbf{40.2\%} \\
        \midrule
    

    \multicolumn{2}{l}{Co-teaching  \cite{Han2018c}} \vline& 89.9  & 87.8  & 87.2  & 87.4  & 62.3  & 40.5   \\
    \multicolumn{2}{l}{JoCoR  \cite{DBLP:conf/cvpr/WeiFC020}}\vline & 90.6  & 88.8  & 88.5  & 88.1  & 66.7  & 40.1   \\
    \multicolumn{2}{l|}{DivideMix \cite{DBLP:conf/iclr/LiSH20}}&93.2 & 92.8 &92.6 & 93.1 & 89.2 & 55.2  \\
    \multicolumn{2}{l}{Co-learning   \cite{DBLP:conf/mm/TanXWL21}}\vline & 92.4  & 91.3  & 91.2  & 91.4  & 81.0  & 47.9   \\
     RoLR \cite{Chen2021TwoWD} & & 95.4 & 94.9 & 94.7 & 95.2 & 92.3 & 62.3 \\
    RankMatch \cite{DBLP:conf/iccv/Zhang0FLCLL23}& & 95.6 & 94.8 & 95.1 & 95.3 & 92.8 & 65.2 \\
    CCL\cite{DBLP:conf/aaai/FanL25} & & {96.4}&	{96.0}&	{95.8}	&{96.1}	&{93.1}	& {65.5}  \\
    ANNE \cite{DBLP:journals/pr/CordeiroC25} & & 96.2 & 95.7 & 95.5 & 95.9 & 93.0 & 66.0 \\
    \midrule
    RoLR + NegScale (Ours) & & \textbf{96.6} & \textbf{96.2} & \textbf{96.0} & \textbf{96.4} & \textbf{93.5} & \textbf{66.3}  \\

    \bottomrule
    \end{tabular}%
    \caption{Comparison with state-of-the-art methods on CIFAR-N. The results are from \cite{Wei2022MitigatingMO} and our replication.}
  \label{tab:main_3}%
\end{table*}%

\begin{table}[htbp]
    \centering
        \small
    \begin{tabular}{cccccc}
    \toprule
            NCT & RoLR & DISC  & CCL & ANNE & Ours\\
         \midrule
            84.1& 88.5 & 87.1 & {89.7} & 88.2 & \textbf{90.7} \\
         \bottomrule
    \end{tabular}
    \caption{Comparison with other methods on Animal-10N. The results of other methods are from \cite{DBLP:journals/pr/CordeiroC25}.}
    \label{tab:main_4}
\end{table}

Following prior works on generalization and robustness under label noise \cite{DBLP:conf/nips/BartlettFT17,DBLP:conf/aistats/HuhR24},
we assume that a linear classifier $W = [W_1, \dots, W_C] \in \mathbb{R}^{d \times C}$ predicts class labels via:
\begin{equation}
  \hat{y}_i = \arg\max_{c} W_c^\top f_i
\end{equation}
and the classifier achieves a clean margin $\gamma_0 > 0$, i.e.,
\begin{equation}
W_{y_i}^\top f_i - \max_{c \ne y_i} W_c^\top f_i \ge \gamma_0.
\end{equation}
\begin{theorem}[Classification Error Bound under NegScale]
\label{the:generalization_error_bound}
Let $f_\theta: \mathcal{X} \rightarrow \mathbb{R}^d$ be a normalized feature extractor, and let $\mathcal{F} = \{f_i = f_\theta(x_i)\}$ denote the feature representations learned on clean labels. Let $\tilde{\mathcal{F}} = \{\tilde{f}_i\}$ denote the feature representations learned under uniform label noise with rate $\tau < \frac{1}{2}$.
The generalization error under label noise using NegScale is bounded by:
\begin{equation}
  |\mathbb{E}\left[ \mathds{1}(\hat{y}_i^{\text{clean}} \ne y_i) \right]-\mathbb{E} \left[ \mathds{1}(\hat{y}_i^{\text{noisy}} \ne y_i) \right]|
\;\le\; \frac{2\epsilon}{\gamma_0}
\end{equation}
where 
$\epsilon = (G \cdot \tau)/(\lambda \cdot \delta_{\text{SNOP}} + \mu \cdot \delta_{\text{DCSA}}$)
.
\end{theorem}


\begin{remark}
\cref{the:generalization_error_bound}  shows that, under small feature perturbations $\epsilon$—which are effectively controlled by  NegScale—the increase in generalization error due to label noise is linearly bounded in $\epsilon/\gamma$. Thus, robustness to label noise is directly linked to the preservation of dissimilarity constraints during training.
\end{remark}


\section{Experiments}


\noindent\textbf{Datasets.} 
To verify the effectiveness of our method, we perform our method on classification tasks with six benchmarks: CIFAR-10 \cite{krizhevsky2009learning}, CIFAR-100 \cite{krizhevsky2009learning}, CIFAR-10N \cite{DBLP:conf/iclr/WeiZ0L0022},  CIFAR-100N \cite{DBLP:conf/iclr/WeiZ0L0022}, Animal-10N \cite{Song2019} and WebVision \cite{DBLP:journals/corr/abs-1708-02862}. The last four benchmarks are real-world noisy datasets.

\noindent\textbf{Implementation Details.}
We conduct experiments using noise types Symmetric (Sym), Asymmetric (Pair), and Instance-dependent (Ins) noise for evaluation.
All reported results are averaged over the last 10 training epochs. For methods lacking available results in original papers, we reimplemented and reproduced them under the same evaluation protocol. The weak and strong data augmentations used follow the settings in \cite{Chen2021TwoWD}.  
For hyperparameters, we set $\lambda=\mu=1,\eta_{\max}=5,\eta_{\min}=3$ as default. 
Since our method is designed as a plug-in, we combine it with RoLR as the default setting throughout experiments.
Further implementation details and descriptions of Noise Injections are provided in the Appendix.





\subsection{Experimental Results}

\begin{table}[htbp]
    \centering
    \small
    \begin{tabular}{l|cc|cc}
    \toprule
      \multirow{2}{*}{\textbf{Method}}   & \multicolumn{2}{c|}{\textbf{WebVision}} & \multicolumn{2}{c}{\textbf{ILSVRC12}}  \\
         & \textbf{top-1} & \textbf{top-5} & \textbf{top-1} & \textbf{top-5}  \\
         \midrule
DSOS &	77.8& 92.0& 74.4& 90.8 \\
DivideMix &	77.3& 91.6& 75.2& 90.8 \\
UNICON &	77.6& 93.4& 75.3& 93.7 \\
RoLR & 	81.8& 94.1& 75.5& 93.8 \\
RankMatch &	79.9 & 93.6 & 77.4 & 94.3 \\
 CCL & {82.3} & {94.6} & {78.2}  & {94.9}\\
 ANNE & 82.0& 94.0&76.8 &92.7\\
         \midrule
         RoLR + NegScale &  \textbf{83.1} & \textbf{94.8} & \textbf{79.4} & \textbf{95.0}  \\
         \bottomrule
    \end{tabular}
    \caption{Comparison with state-of-the-art methods on (mini) WebVision dataset. Numbers denote top-1 (top-5) accuracy (\%) on the WebVision and ImageNet ILSVRC12 validation set. The results of other methods are from \cite{DBLP:conf/iccv/Zhang0FLCLL23,DBLP:journals/pr/CordeiroC25}. }
    \label{tab:main_5}
\end{table}
\subsubsection{Results on CIFAR with Synthetic Noise.}


\begin{figure}[t]
\centering
\includegraphics[width=.6\columnwidth]{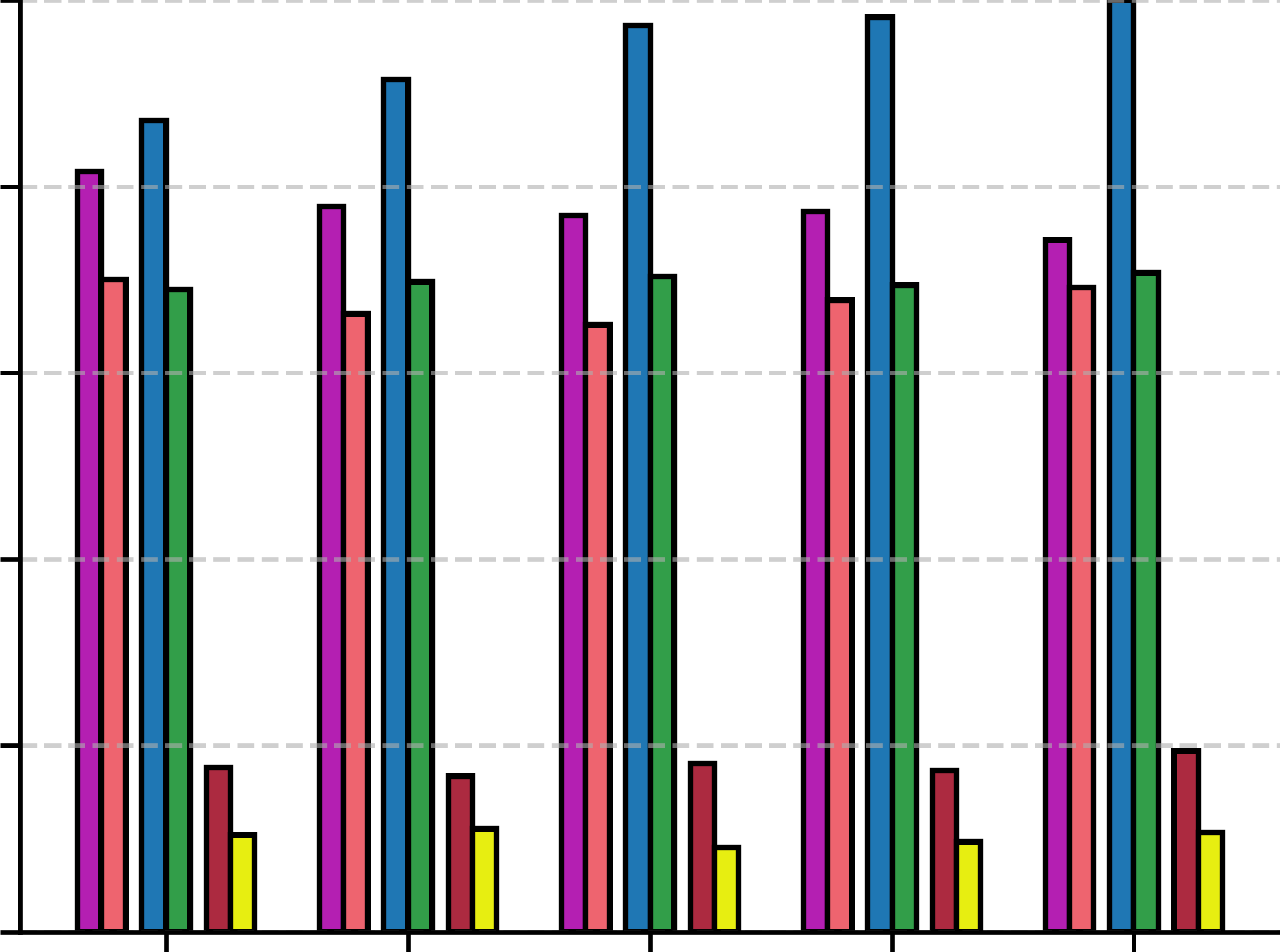} 
\caption{Results on similarity variation after training with NegScale with CIFAR-10 under 50\% symmetric noise. For better visualization, similarity values of \textcolor[HTML]{b41fb2}{\rule{7pt}{7pt}} \textit{Same Class} and \textcolor[HTML]{ee646f}{\rule{7pt}{7pt}} \textit{Same Clean} pairs are divided by 2 to avoid distortion due to their larger magnitudes. The legend is consistent with \cref{fig:app_key_observation}.}
\label{fig:neg_sim}
\end{figure}

\cref{tab:main_1} shows that our plug-in method, when combined with RoLR as the base, consistently outperforms state-of-the-art approaches across all noise levels on both CIFAR-10 and CIFAR-100 under various synthetic noise settings.  
In comparison to sample selection methods such as RankMatch \cite{DBLP:conf/iccv/Zhang0FLCLL23}, our approach achieves a notable performance gain of 3.6\% (70.8\% vs. 67.2\%) on CIFAR-100 with 80\% noise.  
We also surpass DMLP \cite{DBLP:conf/cvpr/TuZLLLWWZ23} and ANNE \cite{DBLP:journals/pr/CordeiroC25}, nearly across all noise settings, with especially significant improvements on the more challenging CIFAR-10 and CIFAR-100 under heavy noise.

\noindent\textbf{Results on Real-world Datasets.}
\cref{tab:main_3}, \cref{tab:main_4}, and \cref{tab:main_5} present results on CIFAR0N, Animal-10N, and WebVision, respectively.  
Our method consistently outperforms all competing approaches across these real-world noisy datasets, demonstrating strong robustness and generalizability.  
Notably, when compared to UNICON \cite{DBLP:conf/cvpr/KarimRRMS22}—a hybrid method that integrates semi-supervised learning with contrastive learning—our method surpasses SOTA by over 3\% in top-1 accuracy on both the mini-WebVision and ILSVRC12 validation sets, while also matching the best top-5 accuracy on these benchmarks.  
These results highlight the effectiveness of NegScale in real-world noisy scenarios.

\begin{table}[htb]
    \centering
    \small
    \begin{tabular}{l|cc|cc}
    \toprule
    \textbf{Dataset} & \multicolumn{2}{c|}{\textbf{CIFAR-10}} & \multicolumn{2}{c}{\textbf{CIFAR-100}} \\
    \midrule
        \textbf{Noise ratio} & \textbf{50\%} & \textbf{80\%} & \textbf{50\%} & \textbf{80\%} \\
        \midrule
        DivideMix & 94.4 &  92.9 & 74.2 & 59.6 \\
        \quad + NegScale & \textbf{96.5} & \textbf{94.6} & \textbf{77.4} & \textbf{70.3}  \\
        \midrule
        RankMatch & 95.4 & 94.2 & 77.6 & 67.2 \\
        \quad + NegScale & \textbf{96.6} & \textbf{95.4} & \textbf{78.1} & \textbf{70.8}  \\
        \midrule
        ANNE & 96.2 & 95.3 & 78.1 & 73.0 \\
        \quad + NegScale & \textbf{96.6} & \textbf{96.2} & \textbf{78.7} & \textbf{73.5}  \\
        \bottomrule
    \end{tabular}
    \caption{Test accuracy of NegScale combined with other methods on CIFAR-10/CIFAR-100 with symmetric noise.}
    \label{tab:exp_plug_in}
\end{table}

\noindent\textbf{Results of Plug-in with Various Method.}
We validate our method’s plug-and-play capability by integrating it with DivideMix, RankMatch, and ANNE. As Table \cref{tab:exp_plug_in} shows, incorporating our dissimilarity-based regularization consistently boosts test accuracy on CIFAR-10 and CIFAR-100 under symmetric noise. These results demonstrate its broad applicability and seamless integration with existing noisy-label learning frameworks.

\noindent\textbf{Results on Similarity.}
\cref{fig:neg_sim} illustrates the changes in inter-sample similarity after training with NegScale. We observe that our method effectively preserves the dissimilarity between semantically unrelated samples, while also mitigating the undesired increase in similarity among semantically related but noisy pairs. This helps reduce the risk of learning spurious semantic correlations introduced by label noise.
Notably, NegScale also increases the similarity between same-class pairs—unlike the degradation observed in \cref{tab:key_observation}—which we attribute to the model learning more robust representations through dissimilarity constraints. This suggests that emphasizing reliable dissimilarity can indirectly facilitate better alignment of truly related samples.


\noindent\textbf{Effects of Components of NegScale.}
We remove the corresponding components to study the effects of each component of our method, such as SNOP and DCSA, and compare them with the full NegScale framework.
As shown in \cref{tab:exp_ab_compents}, removing SNOP or DCSA leads to a significant drop in performance, especially under high noise ratios.  
\begin{table}[htb]
    \centering
    \small
    \begin{tabular}{l|cc|cc}
    \toprule
    \textbf{Dataset} & \multicolumn{2}{c|}{\textbf{CIFAR-10}} & \multicolumn{2}{c}{\textbf{CIFAR-100}} \\
    \midrule
        \textbf{Noise ratio} & \textbf{50\%} & \textbf{80\%} & \textbf{50\%} & \textbf{80\%} \\
        \midrule
        NegScale &  96.6 &  95.6  & 78.7 & 70.8 \\
        w/o SNOP & 95.1  & 94.3 & 76.7 & 67.2\\
        w/o DSCA & 95.3 & 94.8 & 77.0 & 68.1 \\
        w Random Selection & 94.1 & 92.9 & 74.1 & 65.2\\
        \bottomrule
    \end{tabular}
    \caption{Ablation study results of test accuracy (\%) on CIFAR-10 and CIFAR-100 with symmetric  noise.}
    \label{tab:exp_ab_compents}
\end{table}
This indicates that both components are crucial for the effectiveness of our method.  
Additionally, we observe that random selection of negative pairs, which does not leverage the semantic dissimilarity structure, results in a substantial performance degradation, further confirming the importance of structured dissimilarity learning.  
These results validate our design choices and highlight the effectiveness of NegScale in enhancing robustness against label noise.


\begin{figure}
    \centering
    \subfigure[$\lambda$ and $\mu$]
     { \begin{minipage}[t]{0.4\linewidth}
			 \centering
    \includegraphics[width=1\linewidth]{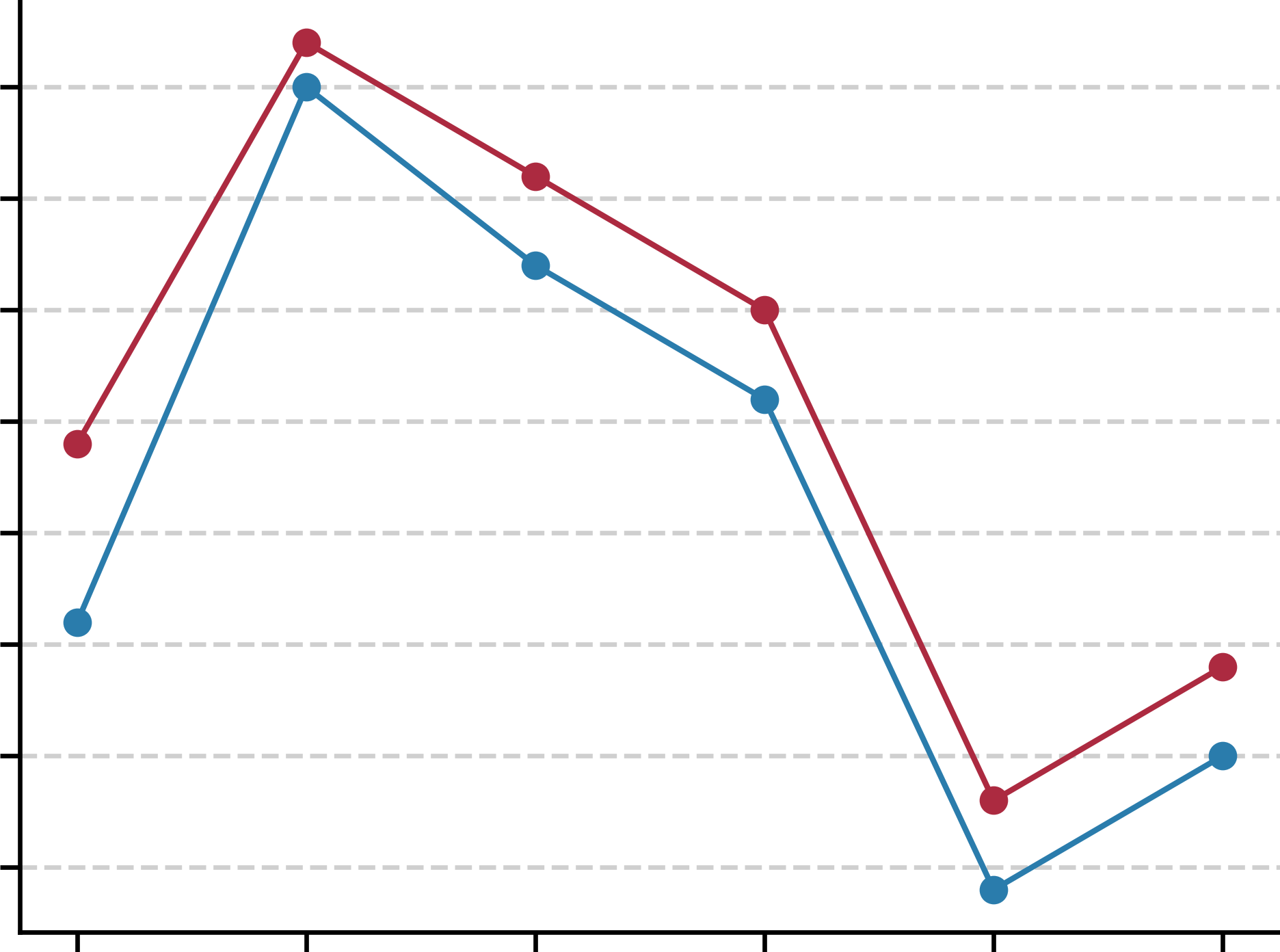}
    \label{fig:sen_lambda_mu}
		\end{minipage}
		}
     \subfigure[$\eta_{\min}$ and $\eta_{\max}$]
     { \begin{minipage}[t]{0.4\linewidth}
			 \centering
    \includegraphics[width=1\linewidth]{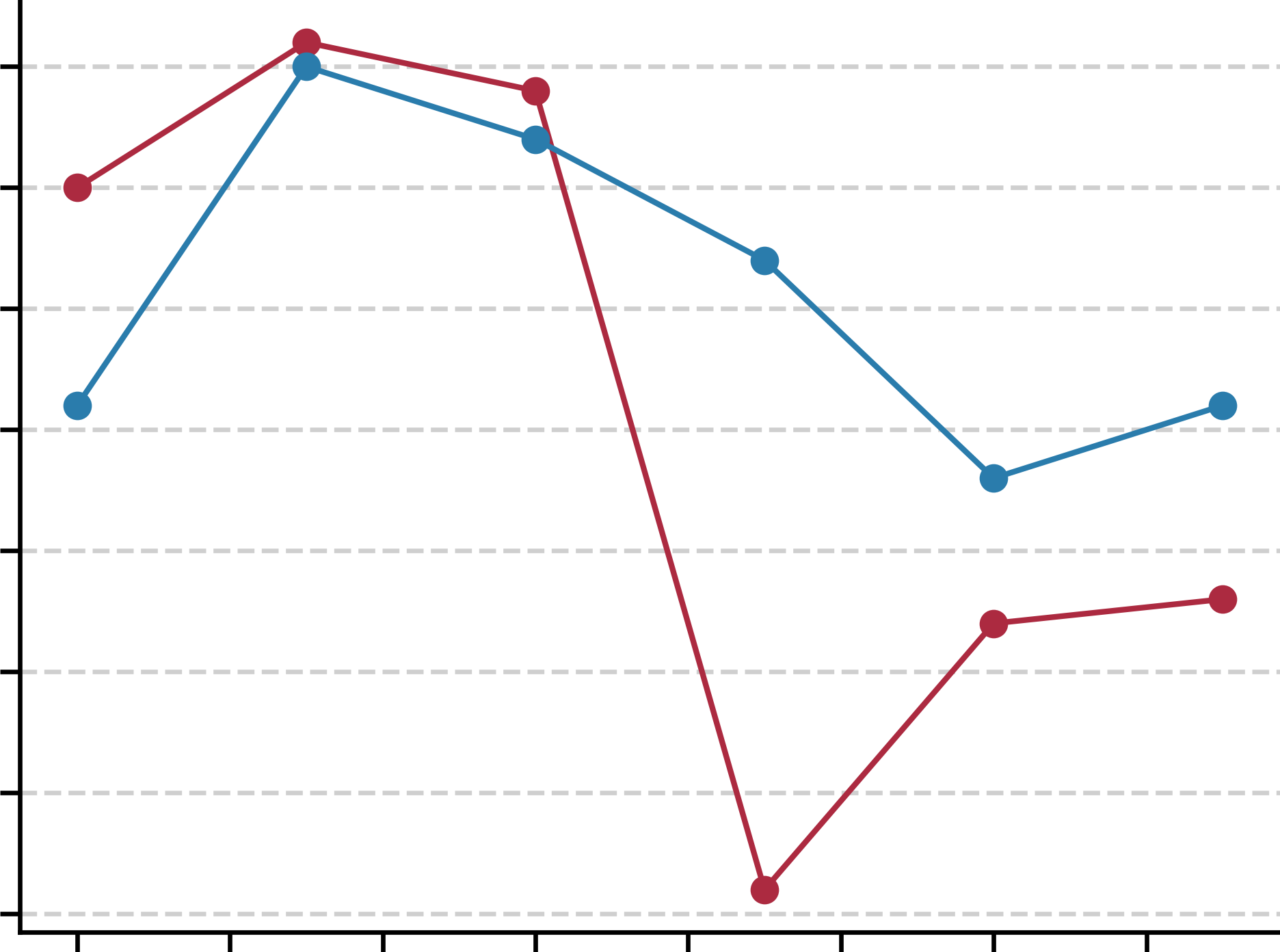}
    \label{fig:sen_tau}
		\end{minipage}
		}
    \caption{Ablation study on hyperparameters with CIFAR-10 under different noises (\textcolor[HTML]{ac2a40}{\rule{7pt}{7pt}} {Sym 0.5}, \textcolor[HTML]{2a7cac}{\rule{7pt}{7pt}} {Ins 0.4}). }
    \label{fig:ab}
\end{figure}

\noindent\textbf{Sensitivity Analysis.}
Our framework introduces four hyperparameters—$\lambda$ and $\mu$ in the total loss (\cref{eq:loss}) and $\eta_{\min},\eta_{\max}$ for negative-pair selection—and we evaluate their impact via sensitivity analysis (see \cref{fig:sen_lambda_mu,fig:sen_tau}). 
In \cref{fig:sen_lambda_mu}, we evaluate different combinations, (0.1, 0.1), (1, 1), (5, 5), (10, 10), (2, 0.5), and (0.5, 2) (from left to right on the x-axis).
We find that setting $\lambda=\mu=1$ yields the best accuracy, demonstrating that a balanced SNOP-DCSA regularization is essential, whereas skewed weights impair robustness. 
In \cref{fig:sen_tau}, we evaluate different combinations, (3,3), (3,5), (5,5), (1,7), (1,3), (4,6), (from left to right on the x-axis).
Similarly, both overly strict and overly loose thresholds $\eta_{\min},\eta_{\max}$ lead to degraded performance, highlighting the necessity of carefully calibrating the dissimilarity range to select informative negative pairs.

\section{Conclusion}


In this paper, we identify and formalize the phenomenon of \textit{Dissimilarity Invariance}, where semantic dissimilarity between unrelated samples remains notably stable even under severe label noise. Motivated by this observation, we propose \textbf{NegScale}, a plug-in framework that explicitly exploits dissimilarity as a robust inductive signal for learning under noisy supervision. 
NegScale consists of two complementary modules: SNOP, which imposes structured orthogonality among negative pairs to enforce local dissimilarity constraints, and DCSA, which calibrates similarity learning by referencing stable dissimilarity across the feature space. 
Extensive experiments on both synthetic and real-world noisy datasets validate the effectiveness of our approach, showing consistent improvements over state-of-the-art baselines. 

\section{Acknowledgments}
This work is supported by Beijing Natural Science Foundation (No.4222037, L181010).

\bibliography{aaai2026}

\end{document}